\documentclass[10pt, conference]{IEEEtran}
\IEEEoverridecommandlockouts

\PassOptionsToPackage{numbers}{natbib} % Use numbered citations

\usepackage[utf8]{inputenc} % allow utf-8 input
\usepackage[T1]{fontenc}    % use 8-bit T1 fonts
\usepackage{hyperref}       % hyperlinks
\usepackage{url}            % simple URL typesetting
\usepackage{booktabs}       % professional-quality tables
\usepackage{amsfonts}       % blackboard math symbols
\usepackage{nicefrac}       % compact symbols for 1/2, etc.
\usepackage{microtype}      % microtypography
\usepackage{xcolor}         % colors
\usepackage{listings}
\usepackage{amsmath}
\usepackage{amssymb}
\usepackage{graphicx}
\usepackage{algorithm}
\usepackage{algpseudocode}
\usepackage{tabularx}
\usepackage{tcolorbox}
\usepackage{multirow}
\usepackage{stfloats} % for positioning of figure* or table* on the same page
\usepackage[utf8]{inputenc}
\title{Bridging LLM Planning Agents and Formal Methods: \\A Case Study in Plan Verification}

\author
{
\IEEEauthorblockN{Keshav Ramani}
\IEEEauthorblockA{
\textit{J.P. Morgan AI Research}\\
New York, USA \\
keshav.ramani@jpmchase.com}
\and
\IEEEauthorblockN{Vali Tawosi}
\IEEEauthorblockA{
\textit{J.P. Morgan AI Research}\\
London, UK \\
vali.tawosi@jpmorgan.com}
\and
\IEEEauthorblockN{Salwa Alamir}
\IEEEauthorblockA{
\textit{J.P. Morgan AI Research}\\
London, UK \\
salwa.alamir@jpmchase.com}
\and
\IEEEauthorblockN{Daniel Borrajo}
\IEEEauthorblockA{
\textit{J.P. Morgan AI Research}\\
Madrid, Spain \\
daniel.borrajo@jpmchase.com}
}

\begin{document}

\maketitle

\begin{abstract}
We introduce a novel framework for evaluating the alignment between natural language plans and their expected behavior by converting them into Kripke structures and Linear Temporal Logic (LTL) using Large Language Models (LLMs) and performing model checking. We systematically evaluate this framework on a simplified version of the PlanBench plan verification dataset and report on metrics like Accuracy, Precision, Recall and F1 scores. Our experiments demonstrate that \mbox{GPT-5} achieves excellent classification performance (F1 score of 96.3\%) while almost always producing syntactically perfect formal representations that can act as guarantees. However, the synthesis of semantically perfect formal models remains an area for future exploration.
\end{abstract}


\section{Introduction}

Large Language Models (LLMs) excel at a wide range of tasks but often lack formal assurances of output correctness. Most evaluations rely on empirical results, without rigorous checks for reliability. We propose a safer design approach: using LLMs to convert natural language into structured formats, then applying classical, deterministic AI methods for reasoning. This leverages LLMs’ strengths in language processing and the reliability of deterministic techniques, especially formal verification, which is widely used in safety-critical systems to provide guarantees.

To integrate these methods, we focus on plan verification. While LLMs are used for planning in areas like agent systems and travel, their outputs lack formal guarantees. Our framework aims to add such guarantees, improving reliability and extending to software verification. Prior work has used symbolic reasoning tools, such as SMT solvers, to verify code \cite{de2010bugs}, and model checking to ensure distributed system correctness \cite{souri2019symbolic}. Our approach advocates translating any language input into a formal representation, then applying verification techniques for robust validation.

In our current work, we explore how translating plans to a formal model with LTL \cite{ltlpaper} specifications can unlock the potential of formal verification. Given the absence of datasets that directly evaluate this setting, we adapt the PlanBench \cite{NEURIPS2023_7a92bcde} plan verification task to align with our objectives. While the original task necessitates identifying specific reasons for plan failures, our approach requires only discerning between valid and invalid plans. Parsing errors encountered during translation are classified as \texttt{unknown}. This work would particularly be of interest to Planning agents, which is an emerging area in Agentic software engineering. Future endeavors may choose to translate to alternative representations, embodying the philosophy that translating to different structured representations can facilitate various tasks, such as using PDDL \cite{mcdermott1998pddl} for planning, among others. The following are our contributions:
\begin{itemize}
    \item \textbf{A Framework for LLM-Driven Formal Verification of plans} that leverages LLMs to translate natural language plans into formal models (Kripke Structures \cite{Kripke1963}) and specifications (LTL), automating model checking and formal verification of plan validity.

    \item \textbf{Empirical Evaluation of LLMs for Plan Verification} through a comprehensive experimental study comparing GPT-4o and GPT-5 on the simplified PlanBench task, reporting accuracy, precision, recall, and F1 for both formal verification and baseline LLM judgment.
\end{itemize}

\begin{figure*}[htbp]
    \centering
    \includegraphics[width=1.00\linewidth, trim={0 0.6cm 0 0},clip]{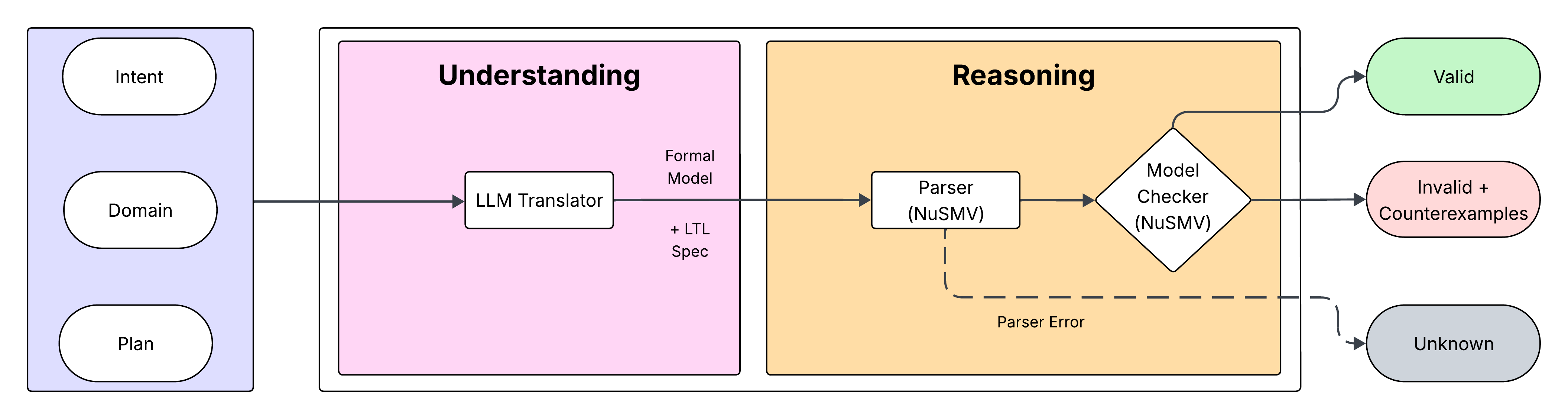}
    \caption{Overview of the LLM-driven plan-specification alignment framework. The Planbench plan verification task contains natural language descriptions of goals, intents, and the environment along with the plan. The LLM addresses the task of understanding such natural language inputs and converting them to a Kripke structure (represented in the NuSMV format) and an LTL specification. For reasoning, we parse this output and provide it to the NuSMV model checker.}
    \label{fig:framework}
\end{figure*}

\section{Related Work}

Multi-agent LLM frameworks like ALMAS \cite{tawosi2025almas} have advanced end-to-end software engineering, but concerns remain about the reliability of LLM-generated code, as highlighted by CodeMirage \cite{agrawal_2024_codemirage}. Automated code assessment using LLMs has also been studied \cite{jensen2023functionality}, emphasizing the need for strong verification methods.

Recent studies have combined LLMs with formal verification and automated planning. VeCoGen \cite{sevenhuijsen2025vecogenautomatinggenerationformally} and Lemur \cite{wu2024lemurintegratinglargelanguage} show LLMs’ potential for automating code generation and verification. Jha et al. \cite{jha2024co} and Hassan et al. \cite{10546729} explore co-synthesis and mutation testing, demonstrating LLMs’ role in guiding formal verification. Our work aligns with these efforts, focusing on plan verification but with broader implications for software verification, while leaving the combined synthesis area for future work.

FVEL \cite{lin2024fvelinteractiveformalverification} and VeriPlan \cite{Lee_2025} integrate LLMs with formal tools for interactive planning. Härer \cite{härer2025specificationevaluationmultiagentllm} and Cemri et al. \cite{cemri2025multiagentllmsystemsfail} address specification and evaluation challenges in multi-agent LLM systems, while Crouse et al. \cite{crouse2024formallyspecifyinghighlevelbehavior} and Tihanyi et al. \cite{tihanyi2025vulnerabilitydetectionformalverification} discuss specification and vulnerability detection. Unlike VeriPlan, our method is simpler, avoids templates, and is evaluated on the PlanBench dataset rather than  user studies.

GenPlanX \cite{borrajo2025genplanxgenerationplansexecution} and "LLMs Can't Plan" \cite{kambhampati2024llmscantplanhelp} critique LLMs’ planning abilities and highlight its challenges. Taxonomies for coordination errors in multi-agent LLMs have also been developed \cite{cemri2025multiagentllmsystemsfail}. Our approach aims to enhance LLM-generated plans with formal guarantees and could extend to automated software verification and agentic software engineering.

\section{Methodology}

Figure \ref{fig:framework} shows an overview of the LLM-driven plan-specification alignment framework. This consists of two key components: Understanding and Reasoning which are both covered in this section in detail.
 
\subsection{Understanding: Natural Language Plans to Formal Model}

Seminal works have reframed the planning problem as a model checking problem \cite{planasmodcheck}. We derive inspiration from these works, specifically utilizing Kripke structures \cite{Kripke1963} to represent plans, similar to how Souri et al. \cite{souri2019symbolic} used it to model the behavior of distributed software systems. By representing plans as Kripke structures, we effectively convert them into state transition systems, where each state corresponds to a unique system configuration, and transitions are modeled as actions that facilitate permissible changes between these configurations. In the context of model checking, a Kripke structure is defined as $ K = (S, S_0, R, L) $, where $ S $ is a finite set of states, each representing a unique system configuration. The set $ S_0 \subseteq S $ contains the initial states from which system executions begin. Transitions $ R $ are defined as actions that facilitate permissible changes between states. The labeling function $ L: S \rightarrow 2^{AP} $ maps each state to a set of atomic propositions from $ AP $, representing properties that hold in those states. 

In order to verify plans, they are represented as sequences of states and actions, where each sequence $ \pi = s_0, s_1, \ldots $ satisfies $ s_0 \in S_0 $ and $ (s_i, s_{i+1}) \in R $ for all $ i \geq 0 $. This approach allows us to analyze and verify the correctness and effectiveness of plans in dynamic environments using model checking techniques. The capability of NuSMV\cite{cimatti1999nusmv} to model these Kripke structures, enables the verification of several formal properties (including goal reachability, safety and liveness), which are specified using Linear Temporal Logic (LTL) with a rich vocabulary of atomic propositions. The syntax of LTL is defined as $$ \varphi ::= p \mid \neg \varphi \mid \varphi_1 \wedge \varphi_2 \mid X\varphi \mid F\varphi \mid G\varphi \mid \varphi_1 U \varphi_2 $$ where $ p \in AP $. LTL semantics are path-based; for instance, $ \pi \models G\varphi $ if $ \varphi $ holds at all positions along $ \pi $. We leverage a large language model (LLM) to generate accurate translations of natural language plans to Kripke Structures in NuSMV and formal properties in LTL. This formulation allows our approach to be applied in a wide variety of dynamic environments, eliminating the traditional dependency on experts to craft specifications on models, as commonly observed in the field of formal verification.

Each aspect of the natural language plan is translated into a formal state-transition system in four key steps. 
\textit{(1) Variables as Representations of Facts: }
In the natural language description, objects and their properties (or facts) are described using terms. The state for each object is captured as a boolean variable. The relationships between objects are recorded via variables (e.g., object \texttt{a} inherits object \texttt{b}, etc).
\textit{(2) Initial Conditions}: The initial conditions described in the natural language plan are encoded into the initial state of the Kripke structure. These assignments are captured by the \texttt{init} command.
\textit{(3) Actions as State Transitions:}
Each action in the natural language plan (e.g., Debug, Refactor, Compile) is described with preconditions and effects. Preconditions are conditions that need to hold prior to actions are included as guards in the conditional updates. Effects are facts becoming true or false after an action are specified by updating the related state variables. 
\textit{(4) Sequencing Using the Stage Variable}
A dedicated variable \texttt{stage} is used to sequence the actions. The natural language plan enumerates a series of actions, and the corresponding stages (e.g., \texttt{s0, s1, ... s10}) ensure that actions are processed in order. Each stage, representing a particular action, triggers its corresponding state transitions and effects.

Finally, the desired outcome is encoded using a LTL specification. In the natural language plan, the goal could be to achieve a set of variables being set to True. In LTL, this is expressed as $F(goal)$ (F symbolizes eventual satisfaction). 

\begin{table*}[t]
    \centering
    \caption{Performance Metrics for the simplified plan verification task from the PlanBench Dataset. One-shot represents the scenario when formal models are generated and checked, w/o FV is when the LLM directly determines plan validity}
    \label{tab:llm_metrics}
    \begin{tabularx}{\textwidth}{ll|XXXXXXXX}
        \toprule
        LLM & Approach & Valid & Invalid & Unk. $\downarrow$ & Accuracy $\uparrow$ & Precision $\uparrow$ & Recall $\uparrow$ & F1 $\uparrow$ & Time\\
        \midrule
        GPT-4o & One-shot & 28.08 & 37.31 & 34.61 & 52.06 & 59.19 & 45.54 & 51.48 & 15.8\\
        GPT-5 & One-shot & 50.64 & 42.50 & \textbf{6.87} & \textbf{95.89} & \textbf{99.44} & \textbf{93.34} & \textbf{96.30} & 47.08\\
        \midrule
        GPT-4o & w/o FV & 40.40 & 59.60 & 0.00 & 80.37 & 97.00 & 67.99 & 79.95 & 7.86\\
        GPT-5 & w/o FV & 57.64 & 42.36 & 0.00 & 99.59 & 99.65 & 99.65 & 99.65 & 15.27\\
        \bottomrule
    \end{tabularx}
\end{table*}

\subsection{Reasoning: Formal Foundations of Model Checking}

One of the variants of model checking that NuSMV implements is known as Bounded Model Checking (BMC) \cite{biere2021bounded}. BMC is a verification technique for finite-state systems that aims to find counterexamples to temporal logic properties within a specified bound $k$ on the execution length. Given a Kripke structure $K$ and a Linear Temporal Logic (LTL) property $\varphi$, BMC translates the search for a counterexample of length $k$ into a propositional satisfiability (SAT) problem. Specifically, BMC constructs a formula $\psi_k$ such that $\psi_k$ is satisfiable if and only if there exists a path $\pi = s_0, s_1, \ldots, s_k$ satisfying the following conditions: $s_0 \in S_0$, $(s_i, s_{i+1}) \in R$ for $0 \leq i < k$, and $\pi \not\models \varphi$. The SAT solver is then employed to determine the satisfiability of $\psi_k$:

\[
    \psi_k := \text{Init}(s_0) \wedge \bigwedge_{i=0}^{k-1} \text{Trans}(s_i, s_{i+1}) \wedge \neg \text{Prop}_\varphi(s_0, \ldots, s_k)
\]

In this formulation, $\text{Init}(s_0)$ encodes the initial states, $\text{Trans}(s_i, s_{i+1})$ encodes the transitions between states, and $\text{Prop}_\varphi(s_0, \ldots, s_k)$ encodes the negation of the property $\varphi$ over the path. If $\psi_k$ is satisfiable, the resulting assignment provides a counterexample of length $k$.

\section{Results}

We evaluate our framework using the simplified PlanBench verification task. For each planning problem, the LLM generates a NuSMV model and LTL property, which are validated using the NuSMV model checker. The output is categorized as \texttt{valid} (SAT), \texttt{invalid} (UNSAT), or \texttt{unknown}. We compare the formally verified output with the baseline LLM judgment to assess the performance trade-offs involved in approximating a formal model of the plan. 

In this study, we explore a single approach to implementing this framework: providing a one-shot example to the LLM to facilitate translation into NuSMV and LTL. We conduct experiments with two LLMs, GPT-4o and GPT-5. GPT-4o is configured with a temperature of 0 to ensure deterministic outputs. For GPT-5  it is to be noted that the temperature parameter is no longer supported, and the reasoning effort parameter is set to 'low'. 

Outputs marked as \texttt{unknown} are excluded from the metric calculations, but are separately reported. The treatment of \texttt{unknown} outputs in evaluation can significantly affect reported metrics and the interpretation of verification results. Counting unknowns as valid can inflate overall accuracy and recall, potentially overstating the system's reliability. Treating unknowns as invalid penalizes the system for each non-adjudicated case, leading to lower accuracy and recall. 
Excluding unknowns from metric calculations offers a clearer assessment of performance on adjudicated cases.

We compare these results against the ground truth labels and report Accuracy, Precision, Recall, and F1-score for each LLM, as presented in Table~\ref{tab:llm_metrics}. The decision to prioritize precision or recall in verification should be guided by the specific application domain and its associated risk profile. 
High precision guarantees reliability, though it may result in reduced coverage (lower recall). Prioritizing recall is appropriate for exploratory, creative, or research-oriented domains where the cost of missing a valid input outweighs the occasional acceptance of an invalid one. High recall enhances coverage but may compromise reliability.

\section{Discussion}

Our experiments confirm the effectiveness of our approach. As shown in Table~\ref{tab:llm_metrics}, GPT-5 achieves high accuracy (95.89\%) and F1 score (96.30), maintaining high performance even while generating formal representations that align with the ground truth regarding the plan's validity. In contrast, GPT-4o’s performance drops sharply, with accuracy and F1 scores around 52\%, mainly due to difficulties in producing correct formal outputs.

Both models occasionally fail to generate syntactically perfect NuSMV models, but GPT-5’s error rate is much lower (6.87\% unknowns in the few-shot setting) than GPT-4o’s (34.61\%). This demonstrates GPT-5’s stronger ability in formal model generation. Our findings highlight the value of formal verification over baseline LLM judgments, which lack formal guarantees. The higher error rate for GPT-4o also suggests that prompt engineering and post-processing are important for improving results.

Initial qualitative analysis shows GPT-5 usually produces syntactically correct NuSMV models and LTL specifications, but further work is needed to ensure semantic accuracy and handle edge cases. In some cases, GPT-5’s models passed verification but did not fully reflect the original plan’s intent, indicating a need for better translation and counterexample analysis.

Overall, our results demonstrate that integrating LLMs with formal verification unlocks new possibilities for reliable, scalable plan validation, while also revealing open challenges in the generation of perfect formal representations that capture planning nuances.

\section{Limitations, Conclusion, and Future Directions}
\label{sec:conclusion}

Our framework improves the reliability and transparency of LLM-generated plans, but several limitations remain. The study reduces PlanBench verification to binary classification and does not yet address the reasons behind plan invalidity or the semantic quality of generated formal models. While we provide empirical and basic qualitative analysis, further work is needed to assess whether LLMs truly capture the intended formal representations. Improving GPT-5’s reasoning and achieving near-perfect accuracy remain open challenges, as does developing a taxonomy of common errors. Our focus is on a simplified verification task, leaving broader applications—such as software and circuit verification—unexplored. There are also concerns about potential misuse, costly errors in critical domains, and bias in generated specifications, underscoring the need for strong safeguards as highlighted in recent studies~\cite{härer2025specificationevaluationmultiagentllm, cemri2025multiagentllmsystemsfail, crouse2024formallyspecifyinghighlevelbehavior}.

In conclusion, our experiments demonstrate that GPT-5 can generate near-perfect NuSMV code and come up with a candidate formal model without significant loss in performance. GPT-4o is inferior to GPT-5 in such tasks, and GPT-5's ability to unlock the area of formal model synthesis can increase the impact of formal verification. By leveraging the strengths of language processing, planning and verification, our framework lays the groundwork for developing robust AI systems capable of sophisticated reasoning and dynamic interaction.

\section*{Disclaimer}
This paper was prepared for informational purposes by the Artificial Intelligence Research group of JPMorgan Chase \& Co and its affiliates (“JP Morgan”), and is not a product of the Research Department of JP Morgan. JP Morgan makes no representation and warranty whatsoever and disclaims all liability, for the completeness, accuracy or reliability of the information contained herein. This document is not intended as investment research or investment advice, or a recommendation, offer or solicitation for the purchase or sale of any security, financial instrument, financial product or service, or to be used in any way for evaluating the merits of participating in any transaction, and shall not constitute a solicitation under any jurisdiction or to any person, if such solicitation under such jurisdiction or to such person would be unlawful.

\bibliographystyle{IEEEtran}
\bibliography{references}

\begin{thebibliography}{10}
\providecommand{\url}[1]{#1}
\csname url@samestyle\endcsname
\providecommand{\newblock}{\relax}
\providecommand{\bibinfo}[2]{#2}
\providecommand{\BIBentrySTDinterwordspacing}{\spaceskip=0pt\relax}
\providecommand{\BIBentryALTinterwordstretchfactor}{4}
\providecommand{\BIBentryALTinterwordspacing}{\spaceskip=\fontdimen2\font plus
\BIBentryALTinterwordstretchfactor\fontdimen3\font minus \fontdimen4\font\relax}
\providecommand{\BIBforeignlanguage}[2]{{%
\expandafter\ifx\csname l@#1\endcsname\relax
\typeout{** WARNING: IEEEtran.bst: No hyphenation pattern has been}%
\typeout{** loaded for the language `#1'. Using the pattern for}%
\typeout{** the default language instead.}%
\else
\language=\csname l@#1\endcsname
\fi
#2}}
\providecommand{\BIBdecl}{\relax}
\BIBdecl

\bibitem{de2010bugs}
L.~De~Moura and N.~Bj{\o}rner, ``Bugs, moles and skeletons: Symbolic reasoning for software development,'' in \emph{International Joint Conference on Automated Reasoning}.\hskip 1em plus 0.5em minus 0.4em\relax Springer, 2010, pp. 400--411.

\bibitem{souri2019symbolic}
A.~Souri, A.~M. Rahmani, N.~J. Navimipour, and R.~Rezaei, ``A symbolic model checking approach in formal verification of distributed systems,'' \emph{Human-centric Computing and Inf. Sciences}, vol.~9, no.~1, p.~4, 2019.

\bibitem{ltlpaper}
A.~Pnueli, ``The temporal logic of programs,'' in \emph{18th Annual Symposium on Foundations of Computer Science (sfcs 1977)}, 1977, pp. 46--57.

\bibitem{NEURIPS2023_7a92bcde}
K.~Valmeekam, M.~Marquez, A.~Olmo, S.~Sreedharan, and S.~Kambhampati, ``Planbench: An extensible benchmark for evaluating large language models on planning and reasoning about change,'' in \emph{Advances in Neural Information Processing Systems}, vol.~36.\hskip 1em plus 0.5em minus 0.4em\relax Curran Associates, Inc., 2023, pp. 38\,975--38\,987.

\bibitem{mcdermott1998pddl}
D.~McDermott, M.~Ghallab, A.~Howe, C.~Knoblock, A.~Ram, M.~Veloso, D.~Weld, and D.~Wilkins, ``Pddl-the planning domain definition language,'' 1998.

\bibitem{Kripke1963}
S.~Kripke, ``Semantical considerations on modal logic,'' \emph{Acta Philosophica Fennica}, vol.~16, pp. 83--94, 1963.

\bibitem{tawosi2025almas}
V.~Tawosi, K.~Ramani, S.~Alamir, and X.~Liu, ``{ALMAS}: an autonomous llm-based multi-agent software engineering framework,'' in \emph{Proceedings of the 1st International Workshop on Multi-Agent Systems using Generative Artificial INtelligence for Automated Software Engineering (MAS-GAIN)}, 2025.

\bibitem{agrawal_2024_codemirage}
\BIBentryALTinterwordspacing
V.~Agarwal, Y.~Pei, S.~Alamir, and X.~Liu, ``Codemirage: Hallucinations in code generated by large language models,'' 2025. [Online]. Available: \url{https://arxiv.org/abs/2408.08333}
\BIBentrySTDinterwordspacing

\bibitem{jensen2023functionality}
R.~I.~T. Jensen, V.~Tawosi, and S.~Alamir, ``Software vulnerability and functionality assessment using llms,'' in \emph{2024 IEEE/ACM International Workshop on Natural Language-Based Software Engineering (NLBSE)}.\hskip 1em plus 0.5em minus 0.4em\relax IEEE, 2024, pp. 25--28.

\bibitem{sevenhuijsen2025vecogenautomatinggenerationformally}
\BIBentryALTinterwordspacing
M.~Sevenhuijsen, K.~Etemadi, and M.~Nyberg, ``Vecogen: Automating generation of formally verified c code with large language models,'' 2025. [Online]. Available: \url{https://arxiv.org/abs/2411.19275}
\BIBentrySTDinterwordspacing

\bibitem{wu2024lemurintegratinglargelanguage}
\BIBentryALTinterwordspacing
H.~Wu, C.~Barrett, and N.~Narodytska, ``Lemur: Integrating large language models in automated program verification,'' 2024. [Online]. Available: \url{https://arxiv.org/abs/2310.04870}
\BIBentrySTDinterwordspacing

\bibitem{jha2024co}
S.~K. Jha, S.~Jha, R.~Ewetz, and A.~Velasquez, ``Co-synthesis of code and formal models using large language models and functors,'' in \emph{MILCOM 2024-2024 IEEE Military Communications Conference (MILCOM)}.\hskip 1em plus 0.5em minus 0.4em\relax IEEE, 2024, pp. 215--220.

\bibitem{10546729}
M.~Hassan, S.~Ahmadi-Pour, K.~Qayyum, C.~K. Jha, and R.~Drechsler, ``Llm-guided formal verification coupled with mutation testing,'' in \emph{2024 Design, Automation \& Test in Europe Conference \& Exhibition (DATE)}, 2024, pp. 1--2.

\bibitem{lin2024fvelinteractiveformalverification}
\BIBentryALTinterwordspacing
X.~Lin, Q.~Cao, Y.~Huang, H.~Wang, J.~Lu, Z.~Liu, L.~Song, and X.~Liang, ``Fvel: Interactive formal verification environment with large language models via theorem proving,'' 2024. [Online]. Available: \url{https://arxiv.org/abs/2406.14408}
\BIBentrySTDinterwordspacing

\bibitem{Lee_2025}
\BIBentryALTinterwordspacing
C.~P. Lee, D.~Porfirio, X.~J. Wang, K.~C. Zhao, and B.~Mutlu, ``Veriplan: Integrating formal verification and llms into end-user planning,'' in \emph{Proceedings of the 2025 CHI Conference on Human Factors in Computing Systems}, ser. CHI ’25.\hskip 1em plus 0.5em minus 0.4em\relax ACM, Apr. 2025, p. 1–19. [Online]. Available: \url{http://dx.doi.org/10.1145/3706598.3714113}
\BIBentrySTDinterwordspacing

\bibitem{härer2025specificationevaluationmultiagentllm}
\BIBentryALTinterwordspacing
F.~Härer, ``Specification and evaluation of multi-agent llm systems -- prototype and cybersecurity applications,'' 2025. [Online]. Available: \url{https://arxiv.org/abs/2506.10467}
\BIBentrySTDinterwordspacing

\bibitem{cemri2025multiagentllmsystemsfail}
\BIBentryALTinterwordspacing
M.~Cemri, M.~Z. Pan, S.~Yang, L.~A. Agrawal, B.~Chopra, R.~Tiwari, K.~Keutzer, A.~Parameswaran, D.~Klein, K.~Ramchandran, M.~Zaharia, J.~E. Gonzalez, and I.~Stoica, ``Why do multi-agent llm systems fail?'' 2025. [Online]. Available: \url{https://arxiv.org/abs/2503.13657}
\BIBentrySTDinterwordspacing

\bibitem{crouse2024formallyspecifyinghighlevelbehavior}
\BIBentryALTinterwordspacing
M.~Crouse, I.~Abdelaziz, R.~Astudillo, K.~Basu, S.~Dan, S.~Kumaravel, A.~Fokoue, P.~Kapanipathi, S.~Roukos, and L.~Lastras, ``Formally specifying the high-level behavior of llm-based agents,'' 2024. [Online]. Available: \url{https://arxiv.org/abs/2310.08535}
\BIBentrySTDinterwordspacing

\bibitem{tihanyi2025vulnerabilitydetectionformalverification}
\BIBentryALTinterwordspacing
N.~Tihanyi, T.~Bisztray, M.~A. Ferrag, B.~Cherif, R.~A. Dubniczky, R.~Jain, and L.~C. Cordeiro, ``Vulnerability detection: From formal verification to large language models and hybrid approaches: A comprehensive overview,'' 2025. [Online]. Available: \url{https://arxiv.org/abs/2503.10784}
\BIBentrySTDinterwordspacing

\bibitem{borrajo2025genplanxgenerationplansexecution}
\BIBentryALTinterwordspacing
D.~Borrajo, G.~Canonaco, T.~de~la Rosa, A.~Garrachón, S.~Gopalakrishnan, S.~Kaur, M.~Morales, S.~Patra, A.~Pozanco, K.~Ramani, C.~Smiley, P.~Totis, and M.~Veloso, ``Genplanx. generation of plans and execution,'' 2025. [Online]. Available: \url{https://arxiv.org/abs/2506.10897}
\BIBentrySTDinterwordspacing

\bibitem{kambhampati2024llmscantplanhelp}
\BIBentryALTinterwordspacing
S.~Kambhampati, K.~Valmeekam, L.~Guan, M.~Verma, K.~Stechly, S.~Bhambri, L.~Saldyt, and A.~Murthy, ``Llms can't plan, but can help planning in llm-modulo frameworks,'' 2024. [Online]. Available: \url{https://arxiv.org/abs/2402.01817}
\BIBentrySTDinterwordspacing

\bibitem{planasmodcheck}
F.~Giunchiglia and P.~Traverso, ``Planning as model checking,'' in \emph{Recent Advances in AI Planning}, S.~Biundo and M.~Fox, Eds.\hskip 1em plus 0.5em minus 0.4em\relax Berlin, Heidelberg: Springer Berlin Heidelberg, 2000, pp. 1--20.

\bibitem{cimatti1999nusmv}
A.~Cimatti, E.~Clarke, F.~Giunchiglia, and M.~Roveri, ``Nusmv: A new symbolic model verifier,'' in \emph{International conference on computer aided verification}.\hskip 1em plus 0.5em minus 0.4em\relax Springer, 1999, pp. 495--499.

\bibitem{biere2021bounded}
A.~Biere, ``Bounded model checking,'' in \emph{Handbook of satisfiability}.\hskip 1em plus 0.5em minus 0.4em\relax IOS press, 2021, pp. 739--764.

\end{thebibliography}

\end{document}